\documentclass{article}
\usepackage{spconf,amsmath,graphicx}
\usepackage[misc]{ifsym}
\usepackage{amssymb}
\usepackage[colorlinks,linkcolor=black,urlcolor=black,citecolor=black]{hyperref}
\usepackage{subfigure}
\usepackage{float}
\usepackage[capitalize]{cleveref}
\usepackage{booktabs}
\usepackage{enumitem}
\usepackage{tabularx}
\usepackage{makecell}
\usepackage{multirow}
\crefname{section}{Sec.}{Secs.}
\Crefname{section}{Section}{Sections}
\crefname{table}{Tab.}{Tabs.}
\Crefname{table}{Table}{Tables}
\crefname{figure}{Fig.}{Figs.}
\Crefname{figure}{Figure}{Figures}
\crefname{equation}{Eq.}{Eqs.}
\Crefname{equation}{Equation.}{Equations.}

\title{GaitGS: Temporal Feature Learning in Granularity and Span Dimension for Gait Recognition}
\name{Haijun Xiong, Yunze Deng, Bin Feng$^{\, \textrm{\Letter}}$, Xinggang Wang, and Wenyu Liu
\thanks{This work was supported by the National Natural Science Foundation of China (No. 62376102).}
\thanks{Corresponding author: Bin Feng (fengbin@hust.edu.cn.)}
}
\address{Huazhong University of Science and Technology, Wuhan, China}
\begin{document}
%
\maketitle
\begin{abstract}
Gait recognition, a growing field in biological recognition technology, utilizes distinct walking patterns for accurate individual identification. However, existing methods lack the incorporation of temporal information. To reach the full potential of gait recognition, we advocate for the consideration of temporal features at varying granularities and spans. This paper introduces a novel framework, {\it GaitGS}, which aggregates temporal features simultaneously in both {\bf granularity} and {\bf span} dimensions. Specifically, the Multi-Granularity Feature Extractor (MGFE) is designed to capture micro-motion and macro-motion information at fine and coarse levels respectively, while the Multi-Span Feature Extractor (MSFE) generates local and global temporal representations. Through extensive experiments on two datasets, our method demonstrates state-of-the-art performance, achieving Rank-1 accuracy of 98.2\%, 96.5\%, and 89.7\% on CASIA-B under different conditions, and 97.6\% on OU-MVLP. The source code will be available at \url{https://github.com/Haijun-Xiong/GaitGS}.
\end{abstract}
\section{Introduction}
\label{sec:intro}

Gait recognition, identifying individuals by analyzing unique walking patterns, surpasses other biometrics like facial, iris, and fingerprint recognition~\cite{sepas2022deep}. Its distinctive advantages include difficulty in disguise and independence from subject cooperation during recognition. Consequently, gait recognition has been widely applied in intelligent security systems, video surveillance, sports science, and crime prevention. However, practical efficacy faces challenges from variations like walking speed and clothing changes, driving ongoing innovation to enhance its effectiveness.

\begin{figure}[t]
    \centering
    \label{fig: motivation}
    \subfigure[The sequence is from subject "079" of CASIA-B. Purple and red boxes depict the movement trends of the head and legs respectively.]{
        \begin{minipage}[l]{0.96\linewidth}
            \centering
            \includegraphics[width=0.75\linewidth]{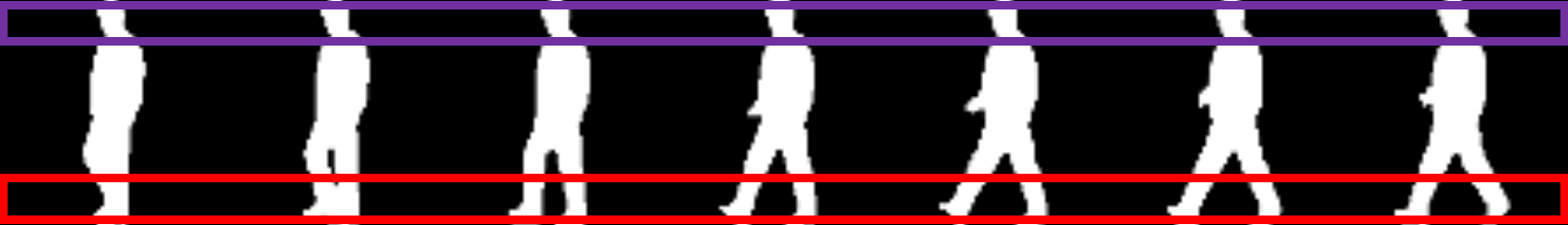}
        \end{minipage}
        \label{fig: granularity}
    }
    \\
    \subfigure[Top to bottom: Sequences from subjects "001", "017" and "063" of CASIA-B, labeled as \uppercase\expandafter{\romannumeral1}, \uppercase\expandafter{\romannumeral2}, and \uppercase\expandafter{\romannumeral3} respectively. Distinguishing between \uppercase\expandafter{\romannumeral1} and \uppercase\expandafter{\romannumeral2} relies on local clues, while \uppercase\expandafter{\romannumeral2} and \uppercase\expandafter{\romannumeral3} require longer time, e.g., global temporal information.]{
        \begin{minipage}[l]{0.96\linewidth}
            \centering
            \includegraphics[width=0.75\linewidth]{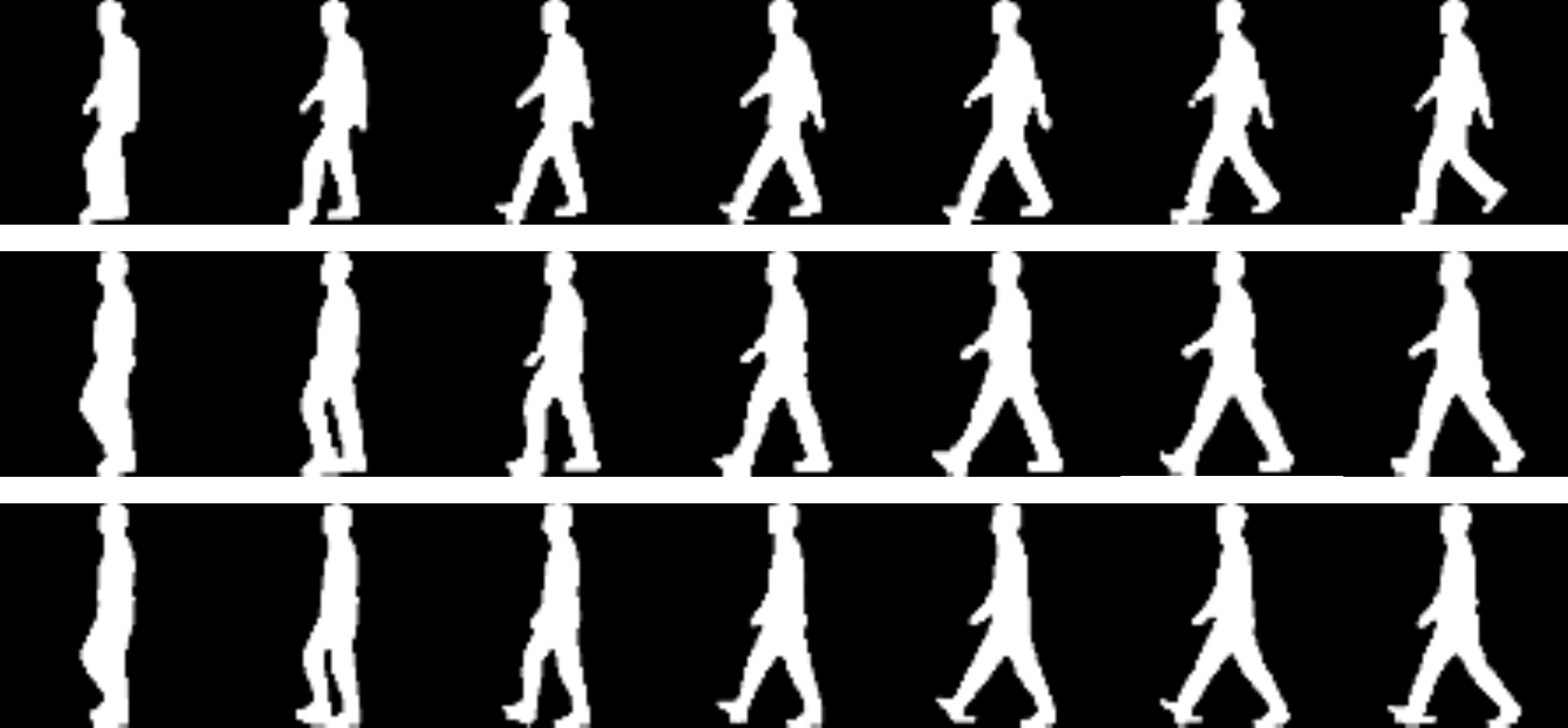}
        \end{minipage}
        \label{fig: span}
    }
    \caption{Motivation: (a) Human body parts display distinct movement patterns during walking, with variations in motion across different regions. (b) Accurate discrimination of gait sequences requires both local and global temporal clues.}
\end{figure}

Currently, in the field of gait recognition, capturing temporal clues is widely acknowledged~\cite{wang2011human, Fan_2020_CVPR, lin2020gait, Lin_2021_ICCV, huang20213d, cui2022GaitTransformer, huang2022star, li2023transgait, Huang_2021_ICCV}. To this end, several noteworthy approaches have emerged in recent research. Wang \textit{et al.}~\cite{wang2011human} divided the gait cycle into multiple phases, creating a Chrono-Gait Image for temporal information incorporation. Fan \textit{et al.}~\cite{Fan_2020_CVPR} introduced Micro-motion Capture Module (MCM) focusing on frame-level temporal feature extraction. Lin \textit{et al.}~\cite{Lin_2021_ICCV} presented Local Temporal Aggregation (LTA) for acquiring local-range temporal information. On the other hand, Cui \textit{et al.}~\cite{cui2022GaitTransformer} and Li \textit{et al.}~\cite{li2023transgait} explored transformer-based methods for global temporal pattern modeling. Huang \textit{et al.}~\cite{Huang_2021_ICCV} proposed CSTL, a temporal modeling network aggregating frame-level, short-term, and long-term temporal features. These methods collectively emphasize the importance of rich and discriminative temporal features within the time series of various granularities and spans.
	
Regarding {\it granularity}, some existing methods use CNN to model temporal information, focusing on fine-granularity information, and capturing details between adjacent frames. However, as shown in \cref{fig: granularity}, extracting useful temporal information within a short timeframe for body parts like the head and torso becomes challenging due to minor changes between adjacent frames. To address this, we propose merging adjacent frames into units to mitigate redundancy and extract temporal information from these units, termed coarse-granularity information. As for {\it span}, the receptive field of CNN is limited, excelling at extracting local temporal information. However, as shown in \cref{fig: span}, there is a clear difference between sequences \uppercase\expandafter{\romannumeral1} and \uppercase\expandafter{\romannumeral2}, while distinguishing sequences \uppercase\expandafter{\romannumeral2} and \uppercase\expandafter{\romannumeral3} solely based on local-span information is a challenge. Therefore, it is crucial to consider longer temporal information, such as global information, which is proven beneficial for gait recognition~\cite{cui2022GaitTransformer}.
	
\begin{figure}[t]
    \centering
    \includegraphics[width=0.9\linewidth]{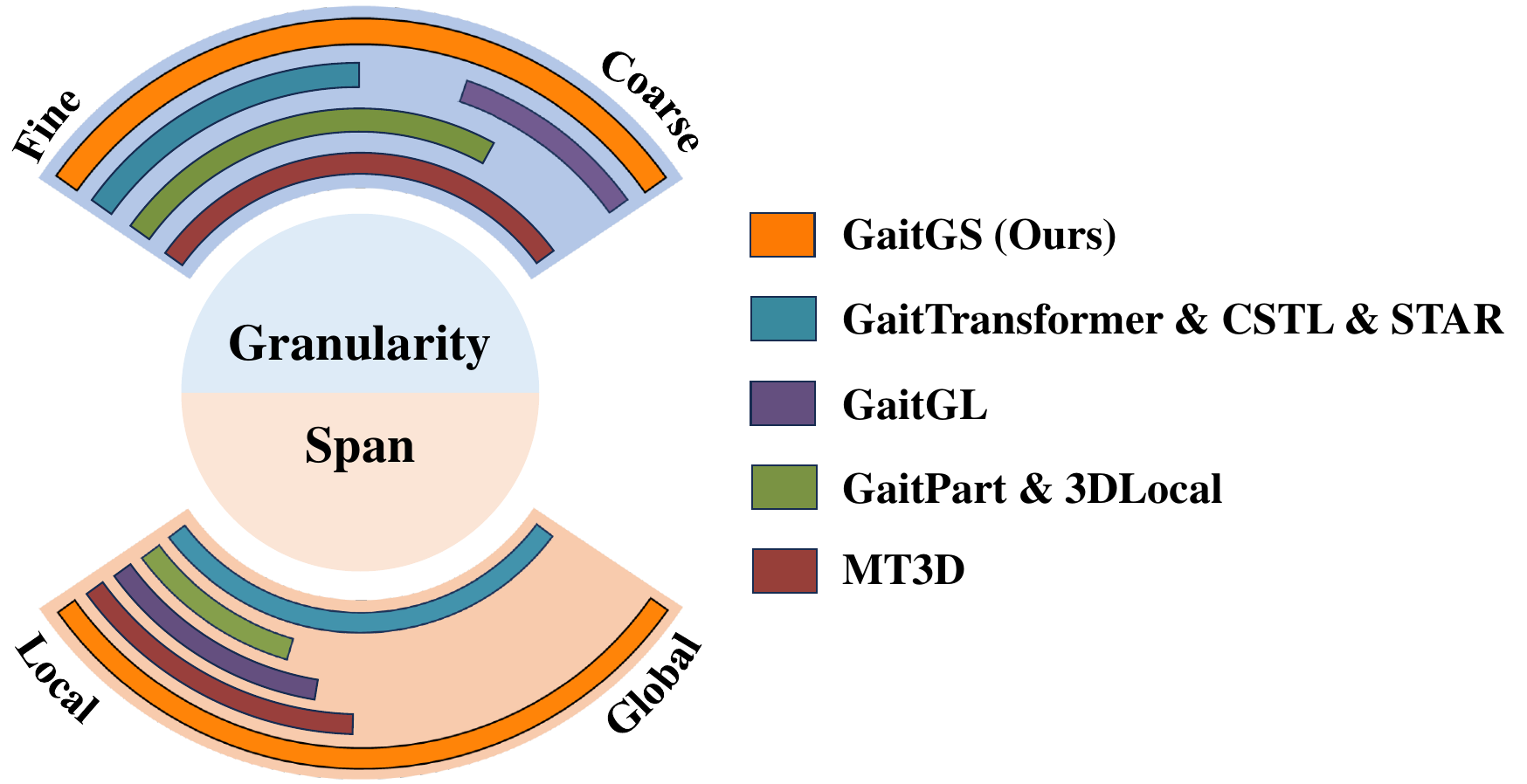}
    \caption{Comparison of temporal modeling methods between previous gait recognition and our method in terms of temporal granularity and span.}
    \label{fig: comparison}
\end{figure}

Building on the above analysis, methods for learning temporal information can be categorized into two main dimensions: granularity and span. In terms of granularity, these methods further divide into fine-level and coarse-level representations. Similarly, in the span dimension, they fall into two distinct types: global and local temporal representations. For example, as depicted in \cref{fig: comparison}, MT3D~\cite{lin2020gait} emphasizes capturing local temporal clues within individual frames, while GaitTransformer~\cite{cui2022GaitTransformer} focuses on fine-level local and global temporal information. However, a notable limitation of current methods is their tendency to model temporal information exclusively from either the granularity or span dimension, resulting in a negative impact on recognition performance.

To overcome this challenge, we introduce an innovative framework, {\bf GaitGS}, for gait recognition. Our approach simultaneously models temporal features from both the granularity and span dimensions. We propose a Multi-Granularity Feature Extractor ({\bf MGFE}) designed for capturing information at various granularity levels. The MGFE includes two CNN-based branches: the Fine Branch Feature Extractor and the Coarse Branch Feature Extractor. Additionally, we present a Multi-Span Feature Extractor ({\bf MSFE}) for local and global temporal modeling. This module operates on both fine and coarse branches, using the MCM~\cite{Fan_2020_CVPR} block for local temporal feature extraction and the Transformer~\cite{vaswani2017attention} block for global temporal feature extraction.

Besides, traditional positional encoding strategies~\cite{vaswani2017attention, dosovitskiy2020vit} have limitations on transformer adaptability. For instance, learnable positional encoding with a fixed length can prove challenging when handling sequences longer than what is observed during training. Meanwhile, sinusoidal encoding may not effectively capture the tiny variations among sequences. To address this, we use grouped convolution to extract temporal position information adaptively, combining the advantages of both strategies above. This results in more accurate position information for global temporal modeling, effectively improving recognition performance.

Our contributions can be summarized as: (1) We introduce GaitGS, a novel and effective framework for gait recognition, simultaneously modeling temporal information from both granularity and span dimensions. (2) The proposed MGFE comprises a fine branch for micro-motion clues and a coarse branch for macro-motion patterns. Additionally, we present MSFE for capturing local and global temporal information. Addressing positional encoding limitations, we integrate grouped convolution to adaptively generate temporal position information, enhancing global temporal modeling. (3) Extensive experiments on CASIA-B~\cite{yu2006framework}, and OU-MVLP~\cite{takemura2018multi} datasets demonstrate the state-of-the-art performance of GaitGS. Furthermore, ablation experiments further confirm the effectiveness of our proposed modules.

\begin{figure*}[t]
    \centering
    \includegraphics[width=0.7\linewidth]{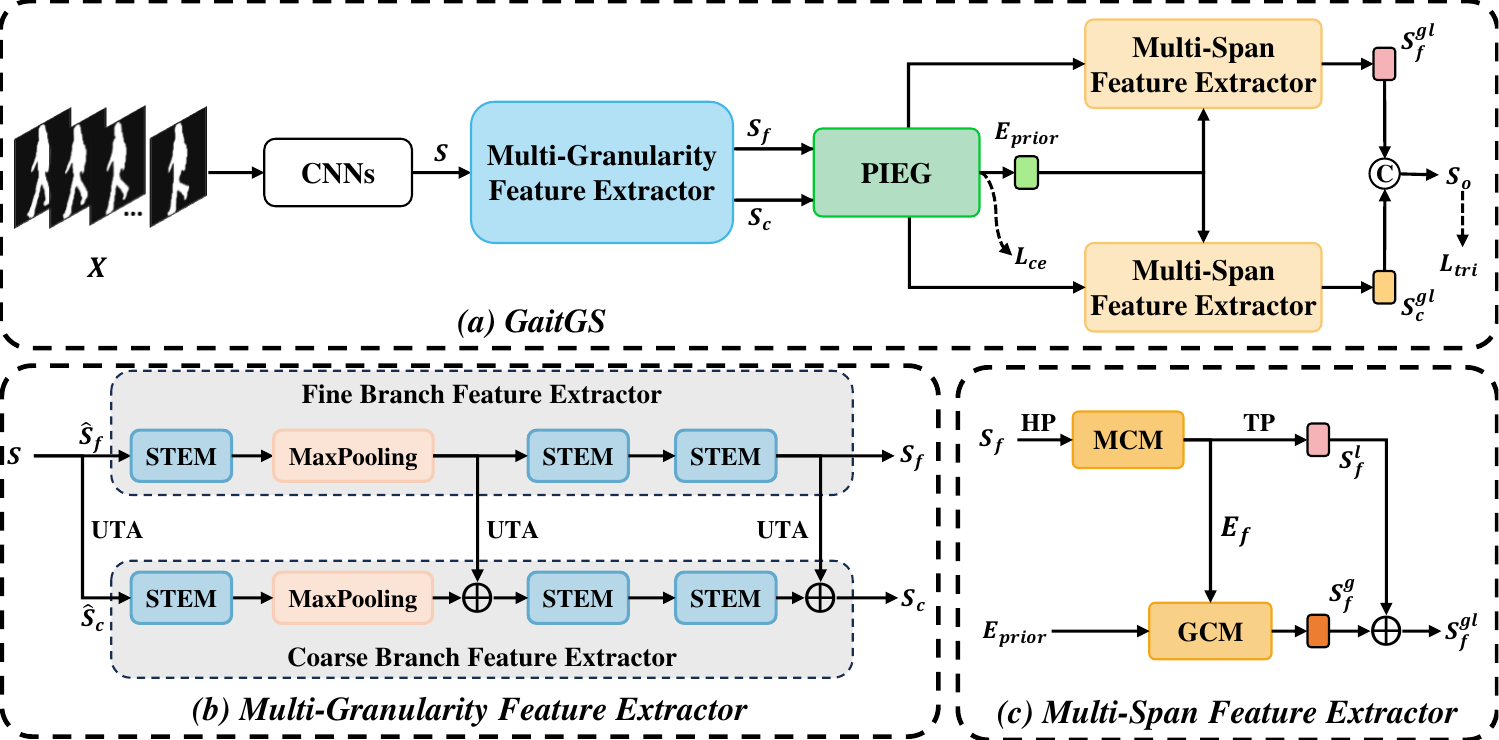}
    \caption{(a) Overview of GaitGS. The Multi-Granularity Feature Extractor (MGFE) extracts both the fine-level feature and coarse-level feature from the initial shallow feature. The Multi-Span Feature Extractor (MSFE) generates local and global temporal information at both fine and coarse levels respectively. (b) Details of MGFE, comprising the Fine Branch Feature Extractor and the Coarse Branch Feature Extractor. The Unit Temporal Aggregation (UTA) operation aims to fuse fine-level information into coarse-level features. (c) Details of MSFE, consisting of MCM~\cite{Fan_2020_CVPR} and Global-information Capture Module (GCM). Take the fine level as an example. MCM captures local temporal details, while GCM focuses on global temporal clues.}
    \label{fig: framework}
\end{figure*}

\section{Related Work}
\label{sec: relatedwork}
{\bf Gait Recognition.} Presently, gait recognition methods fall into two main categories: model-based and appearance-based approaches. {\bf Model-based} methods~\cite{teepe2021gaitgraph, teepe2022towards, zhang2023spatial, pinyoanuntapong2023gaitmixer} analyze human structure to extract body models, including 2D and 3D pose representations, serving as inputs to extract subject identity features. Teepe \textit{et al.}~\cite{teepe2021gaitgraph, teepe2022towards} employed Graph Convolutional Network (GCN) for 2D pose modeling, while Zhang \textit{et al.}~\cite{zhang2023spatial} and Pinyoanuntapong \textit{et al.}~\cite{pinyoanuntapong2023gaitmixer} utilized self-attention mechanism for spatial information extraction from keypoints. On the other hand, {\bf appearance-based} methods~\cite{Fan_2020_CVPR, lin2020gait, Lin_2021_ICCV, huang20213d, cui2022GaitTransformer, huang2022star, Huang_2021_ICCV, han2005individual, chao2019gaitset} learn gait features directly from human morphology, enabling recognition even at low resolutions. Han \textit{et al.}~\cite{han2005individual} synthesized gait sequences into Gait Energy Images (GEI), while Huang \textit{et al.}~\cite{Huang_2021_ICCV} and Lin \textit{et al.}~\cite{Lin_2021_ICCV} used CNNs to extract spatial-temporal features. Compared to model-based methods, appearance-based techniques extract features that tend to involve richer spatial-temporal information.

\noindent{\bf Temporal Modeling.} Recent research~\cite{Fan_2020_CVPR, lin2020gait, Lin_2021_ICCV, huang20213d, cui2022GaitTransformer, huang2022star, Huang_2021_ICCV} in gait recognition has explored various temporal modeling strategies, including 1D CNN, LSTM, 3D CNN, Transformer, and more. Fan \textit{et al.}~\cite{Fan_2020_CVPR} explored local-range motion details within corresponding body parts by parallel MCMs. Lin \textit{et al.}~\cite{Lin_2021_ICCV} utilized 3D CNN to acquire coarse-level temporal features for gait recognition. Huang \textit{et al.}~\cite{Huang_2021_ICCV} introduced a CSTL network dynamically aggregating frame-level, short-term, and long-term temporal features. Cui \textit{et al.}~\cite{cui2022GaitTransformer} proposed the Multiple-Temporal-Scale Transformer, focusing on global-range temporal modeling. However, as shown in \cref{fig: comparison}, these methods primarily emphasize temporal information modeling within either the granularity or span dimension, potentially impacting recognition performance.

\section{Method}
\label{sec: method}
\subsection{Network Pipeline}
\label{sec: pipeline}
As shown in \cref{fig: framework}(a), the dimension of input gait sequence $X$ is denoted as $1 \times T \times H \times W$, where $T$, $H$, and $W$ represent the length, height, and width of the sequence respectively. Initially, a set of 3D CNNs extracts the shallow spatial-temporal feature $S$. Then, the Multi-Granularity Feature Extractor (MGFE) extracts the fine-level feature $S_f$ and coarse-level feature $S_c$. Afterward, the Multi-Span Feature Extractor (MSFE) acquires local and global temporal feature representations in $S_f$ and $S_c$, namely $S_f^{gl}$ and $S_c^{gl}$ respectively. This is achieved by deploying MCM~\cite{Fan_2020_CVPR} and GCM within the fine and coarse branches respectively. Finally, a more robust feature representation $S_o$ is generated by aggregating $S_f^{gl}$ and $S_c^{gl}$ extracted from MSFE. The entire network is trained using a combination of triplet loss and cross-entropy loss to optimize performance.

\subsection{Multi-Granularity Feature Extractor}
As shown in \cref{fig: framework}(b), we propose a dual-branch feature extractor MGFE, designed to capture temporal information at varying granularity levels. The core of MGFE incorporates the Spatial-Temporal Enhanced Module (STEM) based on B3D~\cite{lin2020gait}. To augment the representative capacity of $S_c$, we integrate it with the fine-level feature $S_f$ through UTA.

The STEM first splits the input feature $X_s \in \mathbb{R}^{c \times t \times h \times w}$ into $2^n$ parts along the height dimension, denoted as $\{ X_{s}^{i} | i = 1,2,\cdots,2^n\}$). Then, it utilizes a plain network with $2^n$ B3Ds which share convolutional weights, and a shortcut branch B3D to enhance feature representations. The output of STEM can be expressed as:
\begin{equation}
	X_{SM} = \mathrm{Cat}\{\mathrm{B3D}(X_{s}^{i})\} + \mathrm{B3D}(X_s), i=1,2,\cdots,2^n.
\end{equation}

In each B3D, parallel convolutions with kernel sizes $(3,3,3)$, $(3,1,1)$, and $(1,3,3)$ extract spatial-temporal, pixel-level motion, and salient spatial features, respectively. Afterward, these three features are fused into the final enhanced feature through element-wise addition.

The Feature Extractor, depicted in \cref{fig: framework} (b), comprises the Fine Branch and Coarse Branch Feature Extractors, both using STEM to produce temporal feature representations at different granularities. Fine Branch Feature Extractor involves multiple STEMs and MaxPooling layers, denoted as:
\begin{equation}
	S_{f} = \mathrm{SM}(\mathrm{SM}(\mathrm{MaxPool}^{1\times 2\times 2}(\mathrm{SM}(\hat{S}_f)))).
\end{equation}
Here, $\hat{S}_f \in \mathbb{R}^{C_1\times T \times H \times W}$ and $S_f \in \mathbb{R}^{C_2\times T \times H_2 \times W_2}$, where $S_f$ is the fine-level feature. The function $\mathrm{SM}(\cdot)$ denotes STEM. In the case of Coarse Branch Feature Extractor, the input feature $\hat{S}_c$ is derived from $\hat{S}_f$ through UTA, denoted as:
\begin{equation}
	\hat{S}_c = \mathrm{UTA}(\hat{S}_f),
\end{equation}
where $\hat{S}c \in \mathbb{R}^{C_1 \times T' \times H \times W}$, and the operation $\mathrm{UTA}(\cdot)$ represents the 3D CNN with kernel size $(3, 1, 1)$ and stride $(3, 1, 1)$. The coarse-level feature $S_c \in \mathbb{R}^{C_2\times T' \times H_2 \times W_2}$ is obtained similarly as the fine branch. Additionally, the representation of $S_c$ is enriched by continuously integrating fine-level information. This enables the fine and coarse branches to extract micro-motion and macro-motion information, respectively.

\begin{figure}[t]
    \centering
    \includegraphics[width=0.8\linewidth]{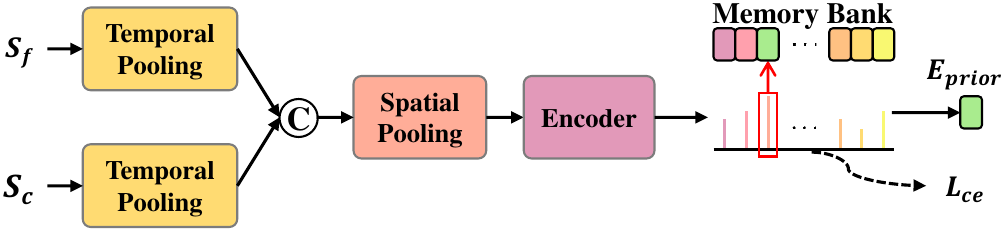}
    \caption{Details of PIEG. The maximum score is marked by the red box, and the selected embedding $E_{prior}$ is indicated by the red arrow.}
    \label{fig: PIEG}
\end{figure}

\subsection{Multi-Span Feature Extractor}
In \cref{fig: framework} (c), MSFE is presented to simultaneously capture local and global temporal clues. This module consists of two key components: MCM and GCM. Before MSFE, the Prior Information Embedding Generation (PIEG) module generates the prior information embedding.

\noindent{\bf Prior Information Embedding Generation.} The PIEG module addresses vulnerability to non-visual variations in fine- and coarse-level features ($S_f$, $S_c$). It extracts the prior information embedding $E_{prior}$ to reduce the influence of gait-unrelated features. As shown in \cref{fig: PIEG}, the aggregation of fine- and coarse-level features is represented as:
\begin{equation}
	S_{prior} = \left(\mathrm{SP}\left(\left(\mathrm{Cat}\left\{\mathrm{TP}(S_{f}), \mathrm{TP}(S_{c})\right\}\right)^p\right)\right)^{1/p},
\end{equation}
where $\mathrm{TP}$ and $\mathrm{SP}$ are temporal and spatial pooling operations, and $p$ is a learnable parameter. The embedding $E_{prior}$ is selected based on the maximum probability $\hat{y}$:
\begin{equation}
	\label{equ: probablity}
	\hat{y} = \arg\max \hat{p} \quad \text{and} \quad \hat{p} = W_{prior} \times S_{prior},
\end{equation}
where $W_{prior} \in \mathbb{R}^{M \times 2C_2}$ is the weight matrix, and cross-entropy loss $L_{ce}$ supervises the probability $\hat{p}$ in \cref{equ: probablity}. Here, $M$ denotes the number of prior information.

\begin{figure}[t]
    \centering
    \includegraphics[width=0.9\linewidth]{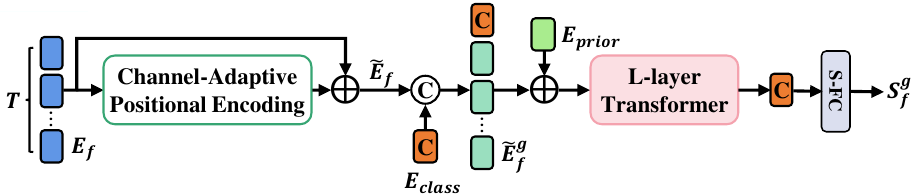}
    \caption{Details of GCM. Taking the generation process of fine-level global temporal feature $S_{f}^g$ as an example.}
    \label{fig: GCM}
\end{figure}

\noindent{\bf Global-information Capture Module.} The GCM is a transformer-based module for extracting global temporal information. \cref{fig: GCM} illustrates the generation process of the global fine-level feature $S_f^g$ as an example. GCM takes two crucial inputs: the prior information embedding $E_{prior}$ and the fine-level feature embedding $E_f$, which is computed as:
\begin{equation}
	E_{f} = \mathrm{MCM}(\mathrm{HP}(S_f)),
\end{equation}
where $\mathrm{MCM}(\cdot)$ and $\mathrm{HP}(\cdot)$ denote the Micro-motion Capture Module and Horizontal Pooling operation. Next, Channel-Adaptive Positional Encoding is introduced to adaptively aggregate adjacent frames for implicit position information. The feature embedding after position embedding is given by:
\begin{equation}
	\tilde{E}_{f} = E_f + \mathrm{GConv}^K_G(E_f)= \{\tilde{E}_{f}^{i} | i = 1, 2, ..., T\},
\end{equation}
where $\mathrm{GConv}^K_G(\cdot)$ is grouped convolution. A learnable class token embedding $E_{class}$ and prior information embedding $E_{prior}$ are introduced, and the final global fine-level feature $S_f^g$ is obtained through an L-Layer Transformer and Separate FC operation. Additionally, the local fine-level feature can be acquired through:
\begin{equation}
	S_f^l=\mathrm{TP}(E_{f}).
\end{equation}

Similarly, global and local coarse-level features are obtained. Finally, features from various granularities and spans are fused to obtain a robust gait representation $S_o$.

\begin{table*}[t]
	\footnotesize
	\caption{Average Rank-1 accuracy (\%) on CASIA-B under all view angles and conditions for input silhouettes with the size of $64 \times 44$ and $128 \times 88$, excluding identical-view case. The best result is indicated in bold, and the second best is underlined.}
	\renewcommand{\arraystretch}{0.9}
	\centering
	\begin{tabular}{c|c|c|ccccccccccc|c}
		\toprule
		\multicolumn{2}{c|}{{\bf Gallery NM}} & \multirow{2}{*}{Resolution}& \multicolumn{11}{c|}{{\bf $0^{\circ}$-$180^{\circ}$}} & \multicolumn{1}{c}{\multirow{2}{*}{Mean}} \\ \cline{1-2} \cline{4-14}
		\multicolumn{2}{c|}{Probe} & &  $0^{\circ}$ & $18^{\circ}$ & $36^{\circ}$ & $54^{\circ}$ & $72^{\circ}$ & $90^{\circ}$ & $108^{\circ}$ & $126^{\circ}$ & $144^{\circ}$ & $162^{\circ}$ & $180^{\circ}$ &  \\ \hline
		\multirow{12}{*}{NM} & GaitSet~\cite{chao2019gaitset} & \multirow{8}{*}{$64\times44$} & 90.8 & 97.9 & 99.4 & 96.9 & 93.6 & 91.7 & 95.0 & 97.8 & 98.9 & 96.8 & 85.8 & 95.0\\
		& GaitPart~\cite{Fan_2020_CVPR} & & 94.1 & 98.6 & 99.3 & 98.5 & 94.0 & 92.3 & 95.9 & 98.4 & 99.2 & 97.8 & 90.4 & 96.2 \\
		& GaitGL~\cite{Lin_2021_ICCV} & & 96.0 & 98.3 & 99.0 & 97.9 & 96.9 & 95.4 & 97.0 & 98.9 & 99.3 & 98.8 & 94.0 & 97.4 \\
		& 3DLocal~\cite{huang20213d} & & 96.0 & 99.0 & {\bf 99.5} & {\bf 98.9} & {\bf 97.1} & 94.2 & 96.3 & {\bf 99.0} & 98.8 & 98.5 & 95.2 & 97.5 \\
		& CSTL~\cite{Huang_2021_ICCV} & & 97.2 & 99.0 & 99.2 & 98.1 & 96.2 & 95.5 & 97.7 & 98.7 & 99.2 & {\bf 98.9} & {\bf 96.5} & 97.8 \\
		& GaitTransformer~\cite{cui2022GaitTransformer} & & 94.9 & 98.3 & 98.4 & 97.8 & 94.8 & 94.1 & 96.3 & 98.5 & 99.0 & 98.3 & 90.7 & 96.5 \\
		& GaitBase~\cite{Fan_2023_CVPR} & & - & - & - & - & - & - & - & - & - & - & - & 97.6  \\
		& {\bf GaitGS (Ours)} & & {\bf 97.5} & {\bf 99.0} & 99.0 & 98.1 & 96.8 & {\bf 96.3} & {\bf 98.1} & 98.7 & {\bf 99.4} & 98.7 & 95.4 & {\bf 97.9} \\ \cline{2-15}
		
		& GLN~\cite{hou2020gait} & \multirow{4}{*}{$128\times88$}  & 93.2 & 99.3 & 99.5 & 98.7 & 96.1 & 95.6 & 97.2 & 98.1 & 99.3 & 98.6 & 90.1 & 96.9 \\
		& 3DLocal~\cite{huang20213d} & & {\bf 97.8} & {\bf 99.4} & {\bf 99.7} & {\bf 99.3} & {\bf 97.5} & 96.0 & {\bf 98.3} & 99.1 & {\bf 99.9} & 99.2 & 94.6 & {\bf 98.3}  \\
		& CSTL~\cite{Huang_2021_ICCV} & & {\bf 97.8} & {\bf 99.4} & 99.2 & 98.4 & 97.3 & 95.2 & 96.7 & 98.9 & 99.4 & 99.3 & {\bf 96.7} & 98.0 \\
		& {\bf GaitGS (Ours)} & & 96.6 & 98.7 & 99.1 & 98.2 & 97.3 & {\bf 97.4} & 98.2 & {\bf 99.5} & 99.4 & {\bf 99.5} & 96.1 & \underline{98.2} \\
		\hline
		
		\multirow{12}{*}{BG} & GaitSet~\cite{chao2019gaitset} & \multirow{8}{*}{$64\times44$} & 83.8 & 91.2 & 91.8 & 88.8 & 83.3 & 81.0 & 84.1 & 90.0 & 92.2 & 94.4 & 79.0 & 87.2 \\
		& GaitPart~\cite{Fan_2020_CVPR} & & 89.1 & 94.8 & 96.7 & 95.1 & 88.3 & 84.9 & 89.0 & 93.5 & 96.1 & 93.8 & 85.8 & 91.5 \\
		& GaitGL~\cite{Lin_2021_ICCV} & & 92.6 & 96.6 & 96.8 & 95.5 & {\bf 93.5} & 89.3 & 92.2 & {\bf 96.5} & {\bf 98.2} & 96.9 & 91.5 & 94.5\\
		& 3DLocal~\cite{huang20213d} & & 92.9 & 95.9 & {\bf 97.8} & {\bf 96.2} & 93.0 & 87.8 & {\bf 92.7} & 96.3 & 97.9 & {\bf 98.0} & 88.5 & 94.3 \\
		& CSTL~\cite{Huang_2021_ICCV} & & 91.7 & 96.5 & 97.0 & 95.4 & 90.9 & 88.0 & 91.5 & 95.8 & 97.0 & 95.5 & 90.3 & 93.6 \\
		& GaitTransformer~\cite{cui2022GaitTransformer} & & 90.3 & 95.9 & 96.0 & 96.0 & 93.1 & 88.1 & 92.2 & 96.1 & 97.5 & 97.5 & 86.1 & 93.5 \\
		& GaitBase~\cite{Fan_2023_CVPR} & & - & - & - & - & - & - & - & - & - & - & - & 94.0  \\
		& {\bf GaitGS (Ours)} & & {\bf 93.9} & {\bf 96.7} & 96.9 & 95.7 & 93.4 & {\bf 89.7} & 92.1 & 95.8 & 97.6 & 97.3 & {\bf 91.6} & {\bf 94.6} \\ \cline{2-15}
		
		& GLN~\cite{hou2020gait} & \multirow{4}{*}{$128\times88$} & 91.1 & 97.7 & 97.8 & 95.2 & 92.5 & 91.2 & 92.4 & 96.0 & 97.5 & 95.0 & 88.1 & 94.0 \\
		& 3DLocal~\cite{huang20213d} & & 94.7 & {\bf 98.7} & {\bf 98.8} & {\bf 97.5} & 93.3 & 91.7 & 92.8 & 96.5 & 98.1 & 97.3 & 90.7 & 95.5  \\
		& CSTL~\cite{Huang_2021_ICCV} & & {\bf 95.0} & 96.8 & 97.9 & 96.0 & 94.0 & 90.5 & 92.5 & 96.8 & 97.9 & {\bf 99.0} & 94.3 & 95.4 \\
		& {\bf GaitGS (Ours)} & & 94.3 & 97.2 & 97.4 & 96.2 & {\bf 96.8} & {\bf 94.5} & {\bf 95.7} & {\bf 98.0} & {\bf 98.7} & 97.7 & {\bf 94.7} & {\bf 96.5} \\
		\hline
		
		\multirow{12}{*}{CL} & GaitSet~\cite{chao2019gaitset} & \multirow{8}{*}{$64\times44$} & 61.4 & 75.4 & 80.7 & 77.3 & 72.1 & 70.1 & 71.5 & 73.5 & 73.5 & 68.4 & 50.0 & 70.4 \\
		& GaitPart~\cite{Fan_2020_CVPR} & & 70.7 & 85.5 & 86.9 & 83.3 & 77.1 & 72.5 & 76.9 & 82.2 & 83.8 & 80.2 & 66.5 & 78.7 \\
		& GaitGL~\cite{Lin_2021_ICCV} & & 76.6 & 90.0 & 90.3 & 87.1 & 84.5 & 79.0 & 84.1 & 87.0 & 87.3 & 84.4 & 69.5 & 83.6 \\
		& 3DLocal~\cite{huang20213d} & & 78.2 & 90.2 & 92.0 & 87.1 & 83.0 & 76.8 & 83.1 & 86.6 & 86.8 & 84.1 & 70.9 & 83.7 \\
		& CSTL~\cite{Huang_2021_ICCV} & & 78.1 & 89.4 & 91.6 & 86.6 & 82.1 & 79.9 & 81.8 & 86.3 & 88.7 & 86.6 & 75.3 & 84.2 \\
		& GaitTransformer~\cite{cui2022GaitTransformer} & & {\bf 81.5} & {\bf 91.9} & 92.2 & {\bf 91.2} & 85.9 & {\bf 83.1} & {\bf 86.8} & {\bf 90.7} & {\bf 90.4} & {\bf 89.0} & 75.6 & {\bf 87.1} \\
		& GaitBase~\cite{Fan_2023_CVPR} & & - & - & - & - & - & - & - & - & - & - & - & 77.4  \\
		& {\bf GaitGS (Ours)} & & 80.4 & 90.9 & {\bf 93.2} & 90.8 & {\bf 86.5} & 83.0 & 85.8 & 89.3 & 90.0 & 88.5 & {\bf 77.0} & \underline{86.9} \\ \cline{2-15}
		
		& GLN~\cite{hou2020gait} & \multirow{4}{*}{$128\times88$} & 70.6 & 82.4 & 85.2 & 82.7 & 79.2 & 76.4 & 76.2 & 78.9 & 77.9 & 78.7 & 64.3 & 77.5 \\
		& 3DLocal~\cite{huang20213d} & & 78.5 & 88.9 & 91.0 & 89.2 & 83.7 & 80.5 & 83.2 & 84.3 & 87.9 & 87.1 & 74.7 & 84.5 \\
		& CSTL~\cite{Huang_2021_ICCV} & & 84.1 & 92.1 & 91.8 & 87.2 & 84.4 & 81.5 & 84.5 & 88.4 & 91.6 & 91.2 & 79.9 & 87.0 \\
		& {\bf GaitGS (Ours)} &  & {\bf 85.3} & {\bf 93.3} & {\bf 93.9} & {\bf 91.1} & {\bf 89.6} & {\bf 85.3} & {\bf 87.5} & {\bf 93.8} & {\bf 93.7} & {\bf 91.9} & {\bf 81.0} & {\bf 89.7} \\ 
		\bottomrule
	\end{tabular}
	\label{tab: CASIA-B}
\end{table*}
\subsection{Loss function}
For efficient network training, we utilize a combination of triplet loss $L_{tri}$ and cross-entropy loss $L_{ce}$. 
The combined loss is given by:
\begin{equation}
	\label{equ: allloss}
	L = L_{tri} + \alpha L_{ce}, 
\end{equation}
where $\alpha$ is cross-entropy weight. 

\section{Experiments}
\label{sec: experiments}
\subsection{Datasets}
{\bf CASIA-B}~\cite{yu2006framework} is a widely used dataset with sequences from 124 subjects, covering various walking conditions and cross-view scenarios. The dataset involves three walking conditions: normal walking (NM, 6 groups), walking with bags (BG, 2 groups), and walking in coats (CL, 2 groups). Each group has 11 different views. We adopt the Large-sample Training (LT) strategy, using subjects \#$001\sim$\#$074$ for training and \#$075\sim$\#$124$ for testing. During testing, the first 4 sequences of NM condition (NM\#1-4) serve as the gallery, and the remaining sequences are split into subsets based on the walking conditions: NM\#5-6 for the NM subset, BG\#1-2 for the BG subset, and CL\#1-2 for the CL subset.

\noindent{\bf OU-MVLP}~\cite{takemura2018multi} is an extensive dataset with cross-view sequences from 10,307 subjects. It consists of two groups, Seq\#$00$ and Seq\#$01$, each having 14 different views. In our experiments, we use 5,153 subjects for training and the remaining subjects for testing. During testing, Seq\#$00$ serves as probe sequences, and Seq\#$01$ serves as gallery sequences.

\begin{table*}[t]
	\footnotesize
	\caption{Average Rank-1 accuracy (\%) on OU-MVLP under 14 probe views excluding invalid probe sequences.}
	\renewcommand{\arraystretch}{0.9}
	\centering
	\begin{tabular}{c|c|c|c|c|c|c|c|c|c|c|c|c|c|c|c}
		\toprule
		\multirow{2}{*}{Method} & \multicolumn{14}{c|}{Probe View} & \multirow{2}{*}{Mean} \\ \cline{2-15}
		& $0^{\circ}$ & $15^{\circ}$ & $30^{\circ}$ & $45^{\circ}$ & $60^{\circ}$ & $75^{\circ}$ & $90^{\circ}$ & $180^{\circ}$ & $195^{\circ}$ & $210^{\circ}$ & $225^{\circ}$ & $240^{\circ}$ & $255^{\circ}$ & $270^{\circ}$ & \\ \hline
		GaitSet~\cite{chao2019gaitset} & 84.5 & 93.3 & 96.7 & 96.6 & 93.5 & 95.3 & 94.2 & 87.0 & 92.5 & 96.0 & 96.0 & 93.0 & 94.3 & 92.7 & 93.3  \\
		GaitPart~\cite{Fan_2020_CVPR} & 88.0 & 94.7 & 97.7 & 97.6 & 95.5 & 96.6 & 96.2 & 90.6 & 94.2 & 97.2 & 97.1 & 95.1 & 96.0 & 95.0 & 95.1 \\
		GLN~\cite{hou2020gait} & 89.3 & 95.8 & 97.9 & 97.8 & 96.0 & 96.7 & 96.1 & 90.7 & 95.3 & 97.7 & 97.5 & 95.7 & 96.2 & 95.3 & 95.6 \\
		GaitGL~\cite{Lin_2021_ICCV} & 90.5 & 96.1 & 98.0 & 98.1 & 97.0 & 97.6 & 97.1 & 94.2 & 94.9 & 97.4 & 97.4 & 95.7 & 96.5 & 95.7 & 96.2 \\
		3DLocal~\cite{huang20213d} & - & - & - & - & - & - & - & - & - & - & - & - & - & - & 96.5   \\
		{\bf GaitGS (Ours)} & {\bf 94.2} & {\bf 97.6} & {\bf 98.8} & {\bf 98.7} & {\bf 97.8} & {\bf 98.5} & {\bf 98.2} & {\bf 96.6} & {\bf 96.9} & {\bf 98.4} & {\bf 98.2} & {\bf 97.3} & {\bf 97.7} & {\bf 97.3} & {\bf 97.6} \\ 
		\bottomrule
	\end{tabular}
	\label{tab: OU-MVLP}
\end{table*}
\subsection{Implementation Details}
{\bf Hyper-parameters.} {\bf 1)} Shallow feature extraction uses an output channel of 32. For CASIA-B, the Fine Branch Feature Extractor has three STEM blocks with output channels set to 64, 64, and 128, and an additional STEM block with an output channel of 256 for OU-MVLP. {\bf 2)} Horizontally pooling divides the feature into 32 parts. {\bf 3)} The number of prior information $M$ is 11 for CASIA-B and for OU-MVLP $M=14$. {\bf 4)} In Channel-Adaptive Positional Encoding, the kernel size $K$ and group number $G$ are set to 7 and 128, respectively. {\bf 5)} The number of transformer blocks $L$ is set to 3 in GCM.

\noindent {\bf Training details.} During training, the Batch ALL sampling strategy is employed with batch sizes of $(8, 8)$ for CASIA-B, and $(32, 8)$ for OU-MVLP. Gait sequences are standardized to a length of 30 frames and a resolution of $64 \times 44$ or $128 \times 88$ pixels. The margin of triplet loss in \cref{equ: allloss} is 0.25, and a weight of $\alpha=0.2$ in \cref{equ: allloss}. Adam is used as the optimizer with weight decay $5\times10^{-4}$ and an initial learning rate of $1\times10^{-4}$. The number of iterations is set to 80K for CASIA-B, and 210K for OU-MVLP. Learning rates drop by $\times0.1$ after 70K iterations for CASIA-B and 150K iterations for OU-MVLP. During testing, the entire sequence is input into the network for gait representations.

\begin{table}[t]
	\centering
	\footnotesize
	\caption{Study of the effectiveness of main components in GaitGS on CASIA-B in terms of the Rank-1 accuracy (\%).}
	\renewcommand{\arraystretch}{0.9}
	\begin{tabular}{c|c|c|c|c}
		\toprule
		\multirow{2}{*}{Method} & \multicolumn{4}{c}{Rank-1 Accuracy} \\ \cline{2-5}
		& NM & BG & CL & Mean \\ \hline
		Baseline & 96.6 & 93.0 & 81.4 & 90.3 \\
		Baseline + MGFE & 97.5 & 94.1 & 84.3 & 91.9 \\
		Baseline + MSFE & 96.8 & 92.0 & 82.9 & 90.6 \\
		Baseline + MSFE + PIEG & 96.9 & 92.4 & 83.2 & 90.8 \\ 
		{\bf Ours} & {\bf 97.9} & {\bf 94.6} & {\bf 86.9} & {\bf 93.1}  \\
		\bottomrule
	\end{tabular}
	\label{tab: ablation-study-1(Components)}
\end{table}

\subsection{Comparison with State-of-the-Art Methods}
{\bf CASIA-B.} 
In \cref{tab: CASIA-B}, our method achieves 97.9\%, 94.6\%, and 86.9\% Rank-1 accuracy in NM, BG, and CL conditions. Compared to other methods, GaitGS performs exceptionally well in NM and BG conditions, with a slightly lower accuracy of 0.2\% than GaitTransformer~\cite{cui2022GaitTransformer} in CL condition. Considering the average Rank-1 accuracy across all conditions, our method excels with 93.1\%, surpassing GaitGL~\cite{Lin_2021_ICCV}, 3DLocal~\cite{huang20213d}, CSTL~\cite{Huang_2021_ICCV} and GaitTransformer~\cite{cui2022GaitTransformer} by 1.3\%, 1.3\%, 1.2\% and 0.7\% respectively. These remarkable results showcase the enhanced feature extraction capabilities of our method and robust temporal modeling mechanism. Furthermore, we evaluate our method using high-resolution images with $128 \times 88$ pixels. These high-resolution images provide richer spatial information, allowing our method to extract high-quality spatial features while efficiently modeling temporal ones. As shown in \cref{tab: CASIA-B}, our method achieves outstanding Rank-1 accuracy of 98.2\%, 96.5\%, and 89.7\% in NM, BG, and CL conditions. Notably, GaitGS outperforms all other methods in BG and CL conditions, achieving the highest average Rank-1 accuracy of 94.8\%. Its temporal modeling excels, especially in parallel and vertical views, where it is less influenced by the absence of spatial information.

\noindent {\bf OU-MVLP.} We extend the evaluation to the extensive OU-MVLP dataset, comparing our method with other competitive approaches. The results presented in \cref{tab: OU-MVLP} show that our method outperforms other competitors on all views and achieves the top average Rank-1 accuracy of an impressive 97.6\% when excluding invalid probe sequences.

\subsection{Ablation Study}
\label{sec: Ablation Study}
{\bf Effectiveness of Main Components.} As detailed in \cref{tab: ablation-study-1(Components)}, ablation experiments are conducted to assess the effectiveness of MGFE, MSFE, and PIEG. Key findings include: 1) MGFE significantly improves performance by 1.6\% compared to the baseline, underscoring its role in enhancing temporal information representations. 2) MSFE alone boosts performance by 0.3\% relative to the baseline, and when coupled with PIEG, the average accuracy increases by 0.5\%, showcasing the contribution of MSFE in capturing local and global temporal clues for richer gait representations. 3) The combined use of all modules yields a substantial 2.8\% improvement in average accuracy compared to the baseline, highlighting the synergies between the modules.

\noindent {\bf Impact of Granularity and Span Dimensional Modeling.} {\it Granularity:} In \cref{tab: ablation-study-2(Fine/Coarse)}, fine and coarse branches show effectiveness individually, improving performance by 1.0\% and 0.6\%, respectively. Simultaneous use and enhancing the coarse branch with UTA yield a 1.3\% average accuracy improvement, highlighting their complementarity in GaitGS. {\it Span:} Results in \cref{tab: ablation-study-3(GCM)} highlight the significance of both local and global temporal modeling. Using global temporal modeling alone shows a decrease in performance compared to local temporal modeling, emphasizing the importance of local information, particularly short-term contextual clues extracted by CNN. The best results come from employing both simultaneously, demonstrating their effectiveness and complementarity.

\begin{table}[t]
    \centering
    \footnotesize
    \caption{Impact of the dual-branch design in terms of the Rank-1 accuracy (\%) on CASIA-B.}
    \renewcommand{\arraystretch}{0.9}
    \begin{tabular}{c|c|c|c|c|c|c}
        \toprule
        \multirow{2}{*}{\makecell{Fine \\ Branch}} & \multirow{2}{*}{\makecell{Coarse \\ Branch}} & \multirow{2}{*}{UTA} & \multicolumn{4}{c}{Rank-1 Accuracy} \\ \cline{4-7}
        &  & & NM & BG & CL & Mean \\ \hline
        & & & 96.9 & 92.4 & 83.2 & 90.8 \\
        $\checkmark$ & & & 97.4 & 93.0 & 85.0 & 91.8 \\
        & $\checkmark$ & & 97.2 & 93.2 & 83.7 & 91.4 \\
        $\checkmark$ & $\checkmark$ & & 97.5 & 94.3 & 85.6 & 92.5 \\
        $\checkmark$ & $\checkmark$ & $\checkmark$& {\bf 97.9} & {\bf 94.6} & {\bf 86.9} & {\bf 93.1}  \\ 
        \bottomrule
    \end{tabular}
    \label{tab: ablation-study-2(Fine/Coarse)}
\end{table}

\begin{table}[t]
	\centering
	\footnotesize
	\caption{Impact of the local and global temporal modeling in terms of the Rank-1 accuracy (\%) on CASIA-B.}
	\renewcommand{\arraystretch}{0.9}
	\begin{tabular}{c|c|c|c|c|c}
		\toprule
		\multirow{2}{*}{Local} & \multirow{2}{*}{Global} & \multicolumn{4}{c}{Rank-1 Accuracy} \\ \cline{3-6} 
		& & NM & BG & CL & Mean \\ \hline
		& & 97.5 & 94.1 & 84.3 & 91.9 \\
		$\checkmark$ & &  97.6 & 94.5 & 85.4 & 92.5 \\
		& $\checkmark$ & 96.2 & 91.9 & 78.7 & 88.9 \\
		$\checkmark$ & $\checkmark$ & {\bf 97.9} & {\bf 94.6} & {\bf 86.9} & {\bf 93.1}  \\ 
		\bottomrule
	\end{tabular}
	\label{tab: ablation-study-3(GCM)}
\end{table}

\section{Conclusion}
\label{sec: conclusion} We introduce GaitGS, a novel gait recognition framework that effectively utilizes temporal information at diverse granularities and spans. The presented MGFE extracts micro-motion and macro-motion clues via fine and coarse branches, while the MSFE dynamically generates positional encoding for comprehensive local and global temporal gait representations. By aggregating temporal features with varying granularities and spans, GaitGS produces robust gait representations, as validated by extensive experiments.

\bibliographystyle{IEEEbib}
{
\bibliography{strings}

\begin{thebibliography}{10}

\bibitem{sepas2022deep}
Alireza Sepas-Moghaddam and Ali Etemad,
\newblock ``Deep gait recognition: A survey,''
\newblock {\em IEEE Transactions on Pattern Analysis and Machine Intelligence}, vol. 45, no. 1, pp. 264--284, 2022.

\bibitem{wang2011human}
Chen Wang, Junping Zhang, Liang Wang, Jian Pu, and Xiaoru Yuan,
\newblock ``Human identification using temporal information preserving gait template,''
\newblock {\em IEEE Transactions on Pattern Analysis and Machine Intelligence}, vol. 34, no. 11, pp. 2164--2176, 2011.

\bibitem{Fan_2020_CVPR}
Chao Fan, Yunjie Peng, Chunshui Cao, Xu~Liu, Saihui Hou, Jiannan Chi, Yongzhen Huang, Qing Li, and Zhiqiang He,
\newblock ``Gaitpart: Temporal part-based model for gait recognition,''
\newblock in {\em Proceedings of the IEEE/CVF conference on computer vision and pattern recognition}, 2020, pp. 14225--14233.

\bibitem{lin2020gait}
Beibei Lin, Shunli Zhang, and Feng Bao,
\newblock ``Gait recognition with multiple-temporal-scale 3d convolutional neural network,''
\newblock in {\em Proceedings of the 28th ACM international conference on multimedia}, 2020, pp. 3054--3062.

\bibitem{Lin_2021_ICCV}
Beibei Lin, Shunli Zhang, and Xin Yu,
\newblock ``Gait recognition via effective global-local feature representation and local temporal aggregation,''
\newblock in {\em Proceedings of the IEEE/CVF International Conference on Computer Vision}, 2021, pp. 14648--14656.

\bibitem{huang20213d}
Zhen Huang, Dixiu Xue, Xu~Shen, Xinmei Tian, Houqiang Li, Jianqiang Huang, and Xian-Sheng Hua,
\newblock ``3d local convolutional neural networks for gait recognition,''
\newblock in {\em Proceedings of the IEEE/CVF International Conference on Computer Vision}, 2021, pp. 14920--14929.

\bibitem{cui2022GaitTransformer}
Yufeng Cui and Yimei Kang,
\newblock ``Gaittransformer: Multiple-temporal-scale transformer for cross-view gait recognition,''
\newblock in {\em 2022 IEEE International Conference on Multimedia and Expo (ICME)}. IEEE, 2022, pp. 1--6.

\bibitem{huang2022star}
Xiaohu Huang, Xinggang Wang, Botao He, Shan He, Wenyu Liu, and Bin Feng,
\newblock ``Star: Spatio-temporal augmented relation network for gait recognition,''
\newblock {\em IEEE Transactions on Biometrics, Behavior, and Identity Science}, vol. 5, no. 1, pp. 115--125, 2022.

\bibitem{li2023transgait}
Guodong Li, Lijun Guo, Rong Zhang, Jiangbo Qian, and Shangce Gao,
\newblock ``Transgait: Multimodal-based gait recognition with set transformer,''
\newblock {\em Applied Intelligence}, vol. 53, no. 2, pp. 1535--1547, 2023.

\bibitem{Huang_2021_ICCV}
Xiaohu Huang, Duowang Zhu, Hao Wang, Xinggang Wang, Bo~Yang, Botao He, Wenyu Liu, and Bin Feng,
\newblock ``Context-sensitive temporal feature learning for gait recognition,''
\newblock in {\em Proceedings of the IEEE/CVF International Conference on Computer Vision}, 2021, pp. 12909--12918.

\bibitem{vaswani2017attention}
Ashish Vaswani, Noam Shazeer, Niki Parmar, Jakob Uszkoreit, Llion Jones, Aidan~N Gomez, {\L}ukasz Kaiser, and Illia Polosukhin,
\newblock ``Attention is all you need,''
\newblock {\em Advances in neural information processing systems}, vol. 30, 2017.

\bibitem{dosovitskiy2020vit}
Alexey Dosovitskiy, Lucas Beyer, Alexander Kolesnikov, Dirk Weissenborn, Xiaohua Zhai, Thomas Unterthiner, Mostafa Dehghani, Matthias Minderer, Georg Heigold, Sylvain Gelly, et~al.,
\newblock ``An image is worth 16x16 words: Transformers for image recognition at scale,''
\newblock {\em arXiv preprint arXiv:2010.11929}, 2020.

\bibitem{yu2006framework}
Shiqi Yu, Daoliang Tan, and Tieniu Tan,
\newblock ``A framework for evaluating the effect of view angle, clothing and carrying condition on gait recognition,''
\newblock in {\em 18th International Conference on Pattern Recognition (ICPR'06)}. IEEE, 2006, vol.~4, pp. 441--444.

\bibitem{takemura2018multi}
Noriko Takemura, Yasushi Makihara, Daigo Muramatsu, Tomio Echigo, and Yasushi Yagi,
\newblock ``Multi-view large population gait dataset and its performance evaluation for cross-view gait recognition,''
\newblock {\em IPSJ Transactions on Computer Vision and Applications}, vol. 10, pp. 1--14, 2018.

\bibitem{teepe2021gaitgraph}
Torben Teepe, Ali Khan, Johannes Gilg, Fabian Herzog, Stefan H{\"o}rmann, and Gerhard Rigoll,
\newblock ``Gaitgraph: Graph convolutional network for skeleton-based gait recognition,''
\newblock in {\em 2021 IEEE International Conference on Image Processing (ICIP)}. IEEE, 2021, pp. 2314--2318.

\bibitem{teepe2022towards}
Torben Teepe, Johannes Gilg, Fabian Herzog, Stefan H{\"o}rmann, and Gerhard Rigoll,
\newblock ``Towards a deeper understanding of skeleton-based gait recognition,''
\newblock in {\em Proceedings of the IEEE/CVF Conference on Computer Vision and Pattern Recognition}, 2022, pp. 1569--1577.

\bibitem{zhang2023spatial}
Cun Zhang, Xing-Peng Chen, Guo-Qiang Han, and Xiang-Jie Liu,
\newblock ``Spatial transformer network on skeleton-based gait recognition,''
\newblock {\em Expert Systems}, p. e13244, 2023.

\bibitem{pinyoanuntapong2023gaitmixer}
Ekkasit Pinyoanuntapong, Ayman Ali, Pu~Wang, Minwoo Lee, and Chen Chen,
\newblock ``Gaitmixer: skeleton-based gait representation learning via wide-spectrum multi-axial mixer,''
\newblock in {\em ICASSP 2023-2023 IEEE International Conference on Acoustics, Speech and Signal Processing (ICASSP)}. IEEE, 2023, pp. 1--5.

\bibitem{han2005individual}
Jinguang Han and Bir Bhanu,
\newblock ``Individual recognition using gait energy image,''
\newblock {\em IEEE Transactions on Pattern Analysis and Machine Intelligence}, vol. 28, no. 2, pp. 316--322, 2005.

\bibitem{chao2019gaitset}
Hanqing Chao, Yiwei He, Junping Zhang, and Jianfeng Feng,
\newblock ``Gaitset: Regarding gait as a set for cross-view gait recognition,''
\newblock in {\em Proceedings of the AAAI conference on artificial intelligence}, 2019, pp. 8126--8133.

\bibitem{Fan_2023_CVPR}
Chao Fan, Junhao Liang, Chuanfu Shen, Saihui Hou, Yongzhen Huang, and Shiqi Yu,
\newblock ``Opengait: Revisiting gait recognition towards better practicality,''
\newblock in {\em Proceedings of the IEEE/CVF Conference on Computer Vision and Pattern Recognition}, 2023, pp. 9707--9716.

\bibitem{hou2020gait}
Saihui Hou, Chunshui Cao, Xu~Liu, and Yongzhen Huang,
\newblock ``Gait lateral network: Learning discriminative and compact representations for gait recognition,''
\newblock in {\em European Conference on Computer Vision}. Springer, 2020, pp. 382--398.

\end{thebibliography}
}

\end{document}